\documentclass[10pt, conference]{IEEEtran}

\IEEEoverridecommandlockouts
\usepackage{cite}
\usepackage{amsmath,amssymb,amsfonts}
\usepackage{algorithmic}
\usepackage{graphicx}
\usepackage{textcomp}
\usepackage{xcolor}
\usepackage{hyperref}
\usepackage{enumitem}
\usepackage{RobStd}

\usepackage{siunitx}
\usepackage{svg}
\usepackage{siunitx}
\usepackage{enumitem}
\usepackage[letterpaper,margin=1in]{geometry}
\def\BibTeX{{\rm B\kern-.05em{\sc i\kern-.025em b}\kern-.08em
    T\kern-.1667em\lower.7ex\hbox{E}\kern-.125emX}}

\begin{document}

\title{LLM-Driven, Datasheet-Aware Automated Hardware Compatibility Verification for Early-Stage, Pre-Schematic Embedded System Design}


\author{
\IEEEauthorblockN{Haotian Qiao and Robert P. Dick}
\IEEEauthorblockA{
Department of Electrical Engineering and Computer Science\\
University of Michigan\\
Ann Arbor, MI, USA\\
qhaotian@umich.edu, dickrp@umich.edu
}
}

\maketitle

\begin{abstract}
In current practice, designers manually inspect datasheets and compare specifications of related components to detect embedded-system design errors before prototyping, a labor-intensive and error-prone process. Large language models (LLMs) provide an opportunity to detect incompatibilities through automated information extraction, semantic reasoning, and code generation. However, straightforward LLM-based ``upload-and-query'' workflows are unreliable for this task for two reasons: long and scattered documentation can lead to property omission, while ad hoc prompts are implicit or ambiguous, resulting in incomplete and inconsistent compatibility verification.


We present an LLM-driven, datasheet-aware framework for early-stage hardware compatibility verification that identifies documentation-level interface incompatibilities based on hardware datasheets and high-level component connectivity descriptions. It does not require, and can therefore be used, before detailed schematic simulation and implementation. We view trustworthy LLM-assisted design automation not as directly generating answers from documents, but as transforming engineering information through traceable verification stages. Given hardware datasheets and high-level component connectivity descriptions, the framework constructs a design graph that captures device connectivity and shared interaction domains, retrieves only the engineering properties required by explicit, domain-oriented verification criteria , and generates deterministic scripts for compatibility evaluation. By decomposing compatibility analysis into modular stages and preserving intermediate results, the framework reduces context overhead, improves transparency and tractability, enables scaling, and avoids reliance on LLMs for numerical computation.

Evaluated on seven embedded-system designs comprising 34 datasheets, our framework achieves 97.5\% compatibility-verification accuracy and an 8.6$\times$ reduction in input context size compared with ``upload-and-query'' workflows. These results demonstrate the feasibility of LLM-assisted, specification-based hardware compatibility verification at an early design stage, as well as the need for, and substantial benefits of, modular task decomposition, formalized verification criteria, and task-aware compact context construction.


\end{abstract}


\section{Introduction}
\label{sec:introduction}
\thispagestyle{plain}

Component incompatibilities may remain hidden until hardware–software integration, where diagnosing their root causes becomes difficult~\cite{menkhaus2005metric, park2020automatic}. Since design defects discovered late in design can incur substantial rework and increase development time and cost~\cite{menkhaus2005metric, van2004defect}, such as PCB respins; industry data reports an average of 2.9 respins per project, with an average cost of \$44,000 per respin~\cite{wiens2018integrated}. Therefore, identification of component incompatibilities before embedded systems proceed to physical implementation or deployment is valuable.

In this paper, hardware compatibility refers specifically to interface-level compatibility, including electrical, communication, power, and operating-environment compatibility required for correct component integration. It does not include software, packaging, or application-specific functional compatibility. Using incompatible hardware or components with conflicting operating environment requirements increases embedded system design effort and cost and may lead to higher in-field failure rates. For example, incompatible voltage levels can prevent reliable communication, supply voltages outside the supported operating range may cause device malfunction or permanent damage, and operating temperatures outside the specified range may lead to unreliable operation or device failure. Undetected hardware incompatibilities may not become apparent until PCB testing or later, where they often require board redesign, component replacement, firmware modification, prototype re-fabrication, and repeated validation. Such design iterations increase engineering cost and extend development schedules by weeks or months. 

This paper presents DEVICES (Datasheet Extraction and Verification of Interface Compatibility for Embedded Systems), an automated LLM-driven framework for interface-level compatibility verification of hardware components based on manufacturer documentation before detailed schematic implementation. DEVICES functions at the early architecture design stage, where designers select system hardware components and define high-level connectivity relationships before developing detailed schematics. Using only hardware datasheets and natural-language connectivity descriptions, DEVICES evaluates whether interconnected components satisfy their documented electrical, communication, and environmental compatibility requirements. Once compatibility is confirmed, designers can proceed to schematic-level design and simulation, which require additional effort to determine pin-level schematics, component configurations, and circuit behavior. Therefore, DEVICES serves as an early-stage screening system that identifies documentation-level incompatibilities before investing the design time for detailed hardware design and implementation.

Consider the following example. A designer of an environmental sensor node PCB can use DEVICES to verify the interface compatibility of hardware components, including the micro-controller, temperature and humidity sensors, power modules, and wireless communication modules, by providing their datasheets and high-level connectivity descriptions. Without such automation, designers must manually search through hundreds of pages of datasheets to identify relevant specifications, a tedious and error-prone process that can take hours for each tentative system-level design. DEVICES automates this process, allowing designers to shift their effort from tedious specification inspection to system design. After compatibility verification succeeds, users can proceed to detailed schematic design for simulation and implementation.

\subsection*{Limitations of Existing LLM-Based ``Upload-and-Query'' Approaches}

Engineers often repeatedly review hardware documentation and perform manual compatibility verification, a process that is labor-intensive, time-consuming, and error-prone. LLMs have demonstrated strong context-based knowledge retrieval, semantic reasoning, and code generation capabilities. By uploading documentation, engineers can use LLMs for tasks such as compatibility evaluation. However, straight-forward application of state-of-the-art LLMs fails at this task. \emph{This paper identifies and overcomes the relevant challenges.}

\textbf{Long and sparse contexts can cause LLMs to overlook distributed engineering information.}
Automated hardware compatibility evaluation requires reasoning over interactions among many components. The straightforward approach is to provide the LLM with the connectivity description and complete datasheets of all involved components. However, this exceeds the effective context lengths of modern LLM services, which typically preprocess uploaded documents and construct query-dependent contexts through document retrieval or chunk selection. Compatibility-related specifications are often scattered across different sections and may not be retrieved together. Even when the required information is included in the model's context, long and sparse contexts can degrade information utilization and instruction following, as demonstrated by prior studies on long-context reasoning~\cite{du-etal-2025-context} and the ``lost-in-the-middle'' phenomenon~\cite{bai2024longbench, liu2024lost}, where LLMs tend to use information less effectively when it appears in the middle of long contexts. 

When multiple datasheets are considered together, the relevant compatibility information is often distributed across a large and heterogeneous context, making it difficult for LLMs to consistently identify all required engineering properties. In our evaluation, five datasheets generated approximately \unit{223}{k} input tokens, yet GPT-5.4 omitted multiple required engineering properties and compatibility specifications even when verification criteria were provided as semantic anchors to guide the identification of target properties; such omissions became more frequent as additional datasheets were introduced. It remains unclear whether future improvements in LLMs will completely eliminate property omission, as both model capabilities and the scale of LLM-based applications continue to increase. This suggests that increasing context capacity alone may be insufficient for reliable knowledge extraction, particularly as the complexity and breadth of engineering tasks grow. We discuss the implications of future LLM evolution for the scalability and long-term value of DEVICES in \autoref{sec: discussion}.

To address this challenge, we construct a compact, task-specific context by retrieving only the engineering properties required by the target verification criteria and organizing related properties closely within the reasoning context. This property-centric context construction reduces irrelevant and redundant information while preserving the information necessary for compatibility analysis, enabling the verification workload to scale with the number of component interactions rather than the amount of datasheet content. 
In the evaluation of the same example involving five datasheets and approximately \unit{223}{k} input tokens, DEVICES uses approximately \unit{18}{k} input tokens across six queries corresponding to different compatibility net nodes (see \autoref{subsec:DG}), with the maximum context size of any individual query remaining below \unit{4.6}{k} tokens.
This focused context allowed the LLM to consistently follow the specified verification procedures and evaluate compatibility based solely on the explicitly provided engineering properties.

\textbf{Ad hoc prompts lack the explicit and complete verification criteria required for reliable compatibility evaluation.}
Reliable compatibility verification requires domain-specific knowledge to identify the complete verification domains, required engineering properties, and evaluation rules for each device interaction or interface, such as SPI, UART, and I$^2$C. Manually providing such detailed specifications is labor-intensive and requires substantial hardware expertise. Prior works~\cite{ma2024death, jin2025understanding, xue2026users} have shown that real-world users predominantly rely on simple, straightforward natural-language prompts, while systematically structured or engineered prompting strategies remain comparatively uncommon.
Consequently, incomplete or ambiguous prompts can cause LLMs to omit required properties or verification procedures, resulting in inconsistent evaluations.

To address this problem, DEVICES formalizes compatibility verification criteria in a structured, domain-oriented way. The required engineering properties, verification procedures, and compatibility rules for each supported interaction or hardware type are defined. The corresponding criteria are automatically selected based on the target connectivity and provided to the LLM, eliminating the need for users to manually construct detailed prompts. This criterion-based design is highly extensible. Supporting a new interaction or hardware type only requires adding its corresponding verification criteria; the underlying framework requires no change. Moreover, because each evaluation query receives only the criteria relevant to the target interaction rather than the complete set of criteria, expanding the supported coverage does not increase the context size of individual evaluations, thereby maintaining verification accuracy as the framework scales.

\textbf{One-shot LLM processing lacks independently verifiable interfaces between engineering evidence and LLM-generated verification operations.}
We view trustworthy LLM-assisted design automation not as directly generating answers from documents, but as transforming engineering information through traceable verification stages. The value of intermediate artifacts is therefore not that they provide LLM reasoning traces, but that they establish controllable and verifiable interfaces between LLM-generated operations and externally stored engineering data.

In DEVICES, the LLM generates structured instructions that reference and operate on externally stored properties rather than repeatedly reproducing engineering information. These instructions can be validated against predefined structures, while the underlying properties remain locally maintained and traceable to hardware knowledge graphs and source datasheets. Each stage therefore passes controlled artifacts to the next stage, allowing both the selected engineering data and the operations applied to them to be independently inspected and verified through deterministic local processing. Although a one-shot LLM can be prompted to generate similar intermediate reasoning, such outputs remain model-generated content and cannot provide the same separation between engineering evidence and LLM-generated operations. Our intermediate artifacts instead constrain information flow across stages, providing traceable evidence for compatibility decisions while preserving the LLM's role in semantic retrieval and verification program generation.

\subsection*{Problem Definition}

Given hardware specifications, such as component datasheets in PDF format, and a natural-language description of the system component connectivity and interactions among components, the objective is to automatically determine whether the interconnected components satisfy the engineering constraints required for their intended integration. Specifically, the evaluation is performed at the interface level and considers the connectivity relationships defined by the high-level connectivity descriptions to identify potential incompatibilities in three dimensions: (1) \emph{electrical and power compatibility}, (2) \emph{communication-interface compatibility}, and (3) \emph{environmental compatibility}. The evaluation is intended for early-stage system design, before schematic capture, simulation, implementation,  deployment, or other prototyping. It relies only on information available from the hardware documentation and high-level connectivity description.

\emph{We view trustworthy LLM-assisted design automation not as directly generating answers from documents, but as transforming engineering information through traceable verification stages. DEVICES decomposes compatibility evaluation into explicit intermediate representations and uses LLMs for controlled reasoning and procedural generation rather than end-to-end decision making. By preserving intermediate artifacts throughout the pipeline, DEVICES enables transparent, auditable, and specification-grounded hardware compatibility verification.}

\subsection*{Design target and scope.}
Our goal is to automate specification-level hardware compatibility verification for embedded systems during early design stages, before detailed schematic capture and simulation. Given hardware specifications, high-level inter-connectivity descriptions, DEVICES models component connectivity and verifies that interconnected components satisfy the relevant electrical, power, communication, and environmental constraints. DEVICES does not predict circuit behavior or replace detailed electronic design automation simulation; instead, it identifies incompatibilities that are explicitly derivable from hardware specifications. Therefore, a successful verification indicates that no incompatibility has been identified within the modeled architecture and specifications.

\subsection*{Terminology}
\label{subsec: terminology}

\textbf{Interface Compatibility} is the condition in which interconnected hardware components satisfy the applicable electrical, communication, and environmental constraints required for their intended interaction, based on their documented specifications and the specified system connectivity.

A \textbf{modular task} is a self-contained compatibility verification subtask that focuses on a specific verification domain and evaluates only the engineering properties and procedures relevant to that domain.

An \textbf{engineering property} is a structured representation of a hardware characteristic that is required for engineering reasoning, verification, or system integration. Engineering properties include quantitative specifications (e.g., supply voltage, maximum clock frequency, power consumption, and operating temperature), and qualitative attributes (e.g., communication protocol, operating mode, and package type), and their associated engineering semantics, such as units, operating conditions, and device variants.

A \textbf{verification domain} defines a category of hardware compatibility that requires a specific set of verification criteria, where each criterion defines the required properties, verification procedure, and compatibility rules for evaluating a specific compatibility requirement. For example, power compatibility and communication interface compatibility are different verification domains because they require different properties and evaluation procedures.

A \textbf{verification criterion} is a structured, domain-specific description of a compatibility verification task that explicitly defines the required engineering properties, the verification procedure, and the deterministic compatibility rule. Each verification criterion corresponds to a particular hardware interaction or compatibility domain (e.g., communication interfaces, power, or environmental conditions) and provides the information required to guide the LLM in constructing and executing the corresponding verification procedure.

\subsection*{Contributions}
\label{sec:contributions}

\begin{tightenumerate}

\item \textbf{Scalable, automated, connectivity-aware LLM-guided interface-level hardware compatibility verification.} We present a connectivity-aware framework that transforms interface-level compatibility verification from implicit natural-language reasoning into an explicit engineering verification procedure. The LLM constructs a design graph from connectivity descriptions to segment component interactions into domain-specific verification tasks, where only relevant verification criteria are evaluated rather than evaluating all compatibility requirements simultaneously. This decomposition enables the verification workload to scale with component interactions rather than the total hardware specification volume. Guided by verification criteria, our framework, DEVICES, retrieves required engineering properties from hardware knowledge graphs and generates deterministic Python verification programs. By integrating hardware knowledge graphs, design graphs, domain-oriented criteria, and deterministic execution, DEVICES enables transparent, reproducible, and scalable verification of power, communication, and environmental compatibility.

\item \textbf{Domain-oriented compatibility verification criteria.} We formalize hardware compatibility verification into a collection of domain-specific verification criteria. Each criterion explicitly specifies the required engineering properties, verification procedure, and deterministic compatibility rules. Rather than relying on implicit or incomplete ad hoc prompting, DEVICES uses the verification criteria associated with each verification domain to generate deterministic verification programs, enabling consistent, reproducible, and extensible compatibility evaluation across different hardware interactions and categories.

\item \textbf{Task-aware context construction through property-specification-centric prompting.} Instead of injecting lengthy raw datasheets into the LLM context, DEVICES constructs the minimal structured context required for each compatibility evaluation by retrieving only the engineering properties specified by the target verification domain. By replacing document-level prompting with property-specification-centric prompting, i.e., associating verification criteria with required properties, DEVICES significantly reduces prompt redundancy, preserves task-specific engineering completeness, and enables scalable, accurate, consistent, and deterministic hardware compatibility evaluation.

\end{tightenumerate}

This paper currently focuses on textual hardware specifications. Image-based specification extraction and interpretation are beyond its scope. Consequently, hardware with specifications primarily provided through figures is not supported in the current implementation. For example, switching regulators, such as buck converters, are not currently modeled because their efficiency is typically characterized using graphs that capture dependencies on multiple operating conditions, including output voltage, load current, and switching frequency, rather than as numerical expressions. Interpreting graphs requires additional figure understanding and operating-point reasoning. Technical diagram interpretation is an active area of research~\cite{lu2022scienceqa, mathew2021docvqa, yue2024mmmu} and such tools can be integrated into our approach.
\section{Related Work}
\label{sec:related works}

Recent work has explored using LLMs for automated hardware design verification by combining design connectivity with manufacturer documentation. Dumont et al.~\cite{dumont2026drcy} describe a multi-agent LLM system for automated schematic connection review. It autonomously retrieves component datasheets, extracts relevant specifications, and performs pin-by-pin analysis to identify semantic connection errors that conventional EDA evaluations cannot detect, such as incorrect pin assignments and regulator configuration errors. Their system conducts schematic-level design review after detailed component connectivity is already available. In contrast, our framework, DEVICES, operates at an earlier, architecture-level design stage and requires only hardware datasheets and high-level system component connectivities described in natural language. Rather than performing open-ended schematic review, we represent the architecture as a design graph of devices and power, data, and environment nets, and evaluate their compatibility according to explicit verification criteria and deterministic constraints.

Pinscope~\cite{pinscope2026} is an automated schematic-review system that combines EDA netlists, bills-of-material and manufacturer datasheets to identify design errors in component specifications. It uses LLM-based datasheet analysis together with deterministic verification to review domains such as power, signal interfaces, pin functions, and electrical constraints across connected components. While both Pinscope and DEVICES perform datasheet-base, netlist-aware hardware verification, they target different stages and abstractions of the design process. Pinscope operates on EDA-level schematic representations with concrete component connectivities, whereas DEVICES operates before detailed schematic implementation using high-level, natural-language component connectivity descriptions. Furthermore, DEVICES explicitly models device interactions as power, data, and environment netnodes and associates each interaction type with verification criteria that define the required engineering properties and evaluation procedures. This formulation enables architecture-level evaluation of power propagation, communication-interface compatibility, and environmental operating-range compatibility before sufficient implementation detail is available for conventional schematic-level verification.

DEVICES relies on an automated hierarchical hardware knowledge graph construction system~\cite{qiao2026faithful} that provides structured and context-aware access to engineering properties extracted from hardware datasheets. Instead of representing specifications as isolated key-value pairs, the knowledge graph preserves the engineering semantics of each property, including associated operating conditions, operation modes, and device variants. Properties are also organized according to a hierarchical hardware ontology to enable efficient retrieval. For example, serial-communication-related properties are grouped under a ``Peripheral'' block, allowing SPI-related specifications to be retrieved through a hierarchy such as Peripheral $\rightarrow$ SPI rather than searching across all extracted properties. The system also performs property deduplication to remove redundant extractions, which commonly arise when the same property is described at multiple locations in a datasheet. Because such occurrences may be physically distant, identifying duplicates through global comparison can be inefficient. Their system instead exploits the observation that properties within the same block share similar engineering semantics and specifications; therefore, deduplication is conducted within each block node, substantially narrowing the deduplication scope and improving the deduplication accuracy. Their system enables DEVICES to construct hierarchical hardware knowledge graphs and retrieve structured compatibility-relevant properties while preserving their engineering context.

The first two related works focus on leveraging datasheet information to assist hardware incompatibility detection based on existing detailed pin-level schematics. In contrast, DEVICES differs from the most advanced closely related work by targeting an earlier architecture-design stage, where detailed pin- and schematic-level connections have not yet been determined. By requiring only hardware datasheets and high-level component connectivity, DEVICES identifies documentation-level incompatibilities before engineers invest additional effort in detailed schematic design and simulation. Therefore, it provides a lightweight early-stage screening mechanism that complements existing lower-level hardware verification systems and traditional EDA verification tools.

\section{Method}

\begin{figure*}[t]
    \centering
    \includegraphics[width=\textwidth]{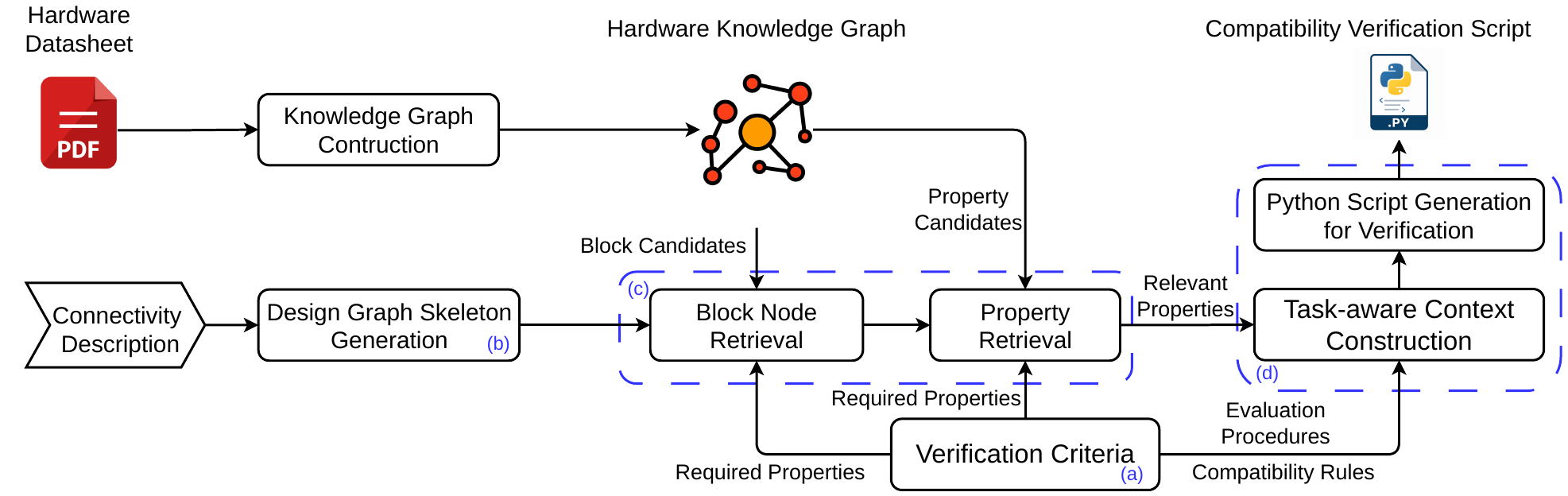}
    \caption{The system pipeline of DEVICES. Verification criteria ((a), \autoref{subsec:formalized verfication criteria}) guide property retrieval and verification script generation. Given a high-level component connectivity description, DEVICES constructs a design graph skeleton to represent system connectivity ((b), \autoref{subsec:DG}). For each netnode, relevant properties are then hierarchically retrieved from the hardware knowledge graph according to the required properties specified by the verification criteria ((c), \autoref{subsec:property retrieval}). Finally, the retrieved properties, corresponding evaluation procedures, and deterministic compatibility rules are assembled into a task-aware context for each netnode, from which DEVICES generates a Python script for compatibility verification ((d), \autoref{subsec: verification script generation and constraint evaluation}). Hardware knowledge graphs are constructed using our companion work.}
    \label{fig:DG-pipeline}
\end{figure*}

Scalability remains a fundamental challenge for LLM-guided hardware compatibility verification because increasing model capability does not eliminate the growth of engineering problem complexity. As LLMs enable the analysis of larger and more sophisticated hardware systems, directly reasoning over the complete specification space becomes impractical. DEVICES addresses this challenge by decomposing compatibility verification into connectivity-aware and domain-oriented modular tasks and constructing compact contexts containing only the properties required for each verification task. A modular task is a self-contained compatibility verification subtask that focuses on a specific verification domain and evaluates only the engineering properties relevant to that domain.
This architecture enables compatibility reasoning to scale with relevant component interactions rather than the entire hardware knowledge space, while allowing future improvements in LLM capabilities to further enhance each modular reasoning task.

\autoref{fig:DG-pipeline} shows the system pipeline of DEVICES.
The LLM constructs a design graph from connectivity descriptions to segment component interactions into domain-specific verification tasks. Driven by verification criteria, it hierarchically retrieves the engineering properties required by the applicable verification criteria from hardware knowledge graphs and generates scripts to execute the specified compatibility verifications. This section first introduces the verification criteria and then describes the three major stages of DEVICES: connectivity-aware design graph construction, hierarchical property retrieval, and verification-criterion-driven evaluation script generation.

\subsection{Verification Criteria}

\label{subsec:formalized verfication criteria}
As discussed in \autoref{sec:contributions}, ad hoc prompt construction does not provide sufficiently precise and complete instructions for compatibility evaluation, particularly regarding what should be verified, what engineering properties are required, and how the properties should be evaluated. Accordingly, it produced inconsistent or incomplete evaluation results. We therefore specify compatibility verification criteria (Component (a) in \autoref{fig:DG-pipeline}) based on domain knowledge and encode them into prompting templates that guide the LLM throughout property retrieval and evaluation. Each verification criterion explicitly specifies evaluation procedure, required engineering properties, and the compatibility rule.

\subsubsection{Verification Domains and Evaluation Procedures}

For each verification domain, the verification criteria provide explicit evaluation procedures, including the required comparisons, mathematical relationships, and fallback rules for missing specifications. For example, for SPI voltage-level compatibility, a criterion specifies that the minimum output-high voltage ($V_{\mathrm{OH},\min}$) of the output device must be greater than or equal to the minimum input-high voltage ($V_{\mathrm{IH},\min}$) of the input device. It further requires that the maximum output-high voltage ($V_{\mathrm{OH},\max}$) be less than or equal to the maximum input-high voltage ($V_{\mathrm{IH},\max}$). When these values are not explicitly specified, the criterion defines deterministic fallback rules: if $V_{\mathrm{OH},\max}$ is unavailable, the typical I/O rail supply voltage is used; if $V_{\mathrm{IH},\max}$ is unavailable, the upper bound of the operating input-pin voltage is preferred, with the corresponding absolute-maximum input voltage used as a fallback.

These explicit rules transform engineering practices that are often applied implicitly by designers into reproducible evaluation procedures that can be consistently followed by the LLM and subsequently executed through generated evaluation scripts.

\subsubsection{Required Engineering Properties}

Each verification criterion also enumerates the engineering properties required to perform its evaluation. Property descriptions are provided to guide the LLM in retrieving the corresponding properties from the hardware knowledge graphs and associating them with the appropriate requirements in the design graph. For the SPI voltage-level criterion above, the required properties include ``$V_{\mathrm{OH}}$ range,'' ``$V_{\mathrm{IH}}$ range,'' ``$V_{\mathrm{DDIO}}$,'' and ``Absolute Maximum Ratings.'' Each property is accompanied by a semantic description to disambiguate the intended engineering information; for example, ``$V_{\mathrm{OH}}$ range'' is defined as the minimum and maximum output voltages associated with the device's output-high logic level.

The explicit enumeration of required properties also allows DEVICES to construct task-specific contexts. During hierarchical property retrieval, the LLM retrieves the properties corresponding to the requirements defined by each applicable verification criterion and associates them with the relevant devices and interactions in the design graph. During evaluation script generation, the required properties are directly provided to the LLM according to the corresponding property requirements, eliminating errors from using the LLM to determine which retrieved properties should be used for each verification.

\subsubsection{Interface- and Device-Type-Dependent Criteria}

The abstraction level at which verification criteria are defined depends on the type of compatibility being evaluated. Communication interfaces provide a natural abstraction because devices implementing the same interface generally share common electrical and protocol-level compatibility requirements. This allows communication verification criteria to be defined at the interface level and reused across devices supporting the same interface, such as SPI, UART, and I$^2$C.

Power compatibility, in contrast, depends on device type because different power-conversion architectures impose different engineering constraints. For example, a low-dropout regulator (LDO) requires input-voltage evaluation that accounts for dropout voltage, and its power-transfer efficiency can be approximated from the input and output voltages. In contrast, the efficiency of switching regulators, such as Buck converters, depends on multiple operating conditions, including output voltage, load current, and switching frequency. Consequently, power verification criteria are defined with respect to device types, whereas communication verification criteria are defined with respect to interfaces. This distinction is incorporated into the verification criterion design to ensure that each compatibility verification is evaluated at the appropriate level of abstraction.

\subsection{Design Graph Construction}

\label{subsec:DG}

\begin{figure}[!t]
    \centering
    \includegraphics[width=\linewidth]{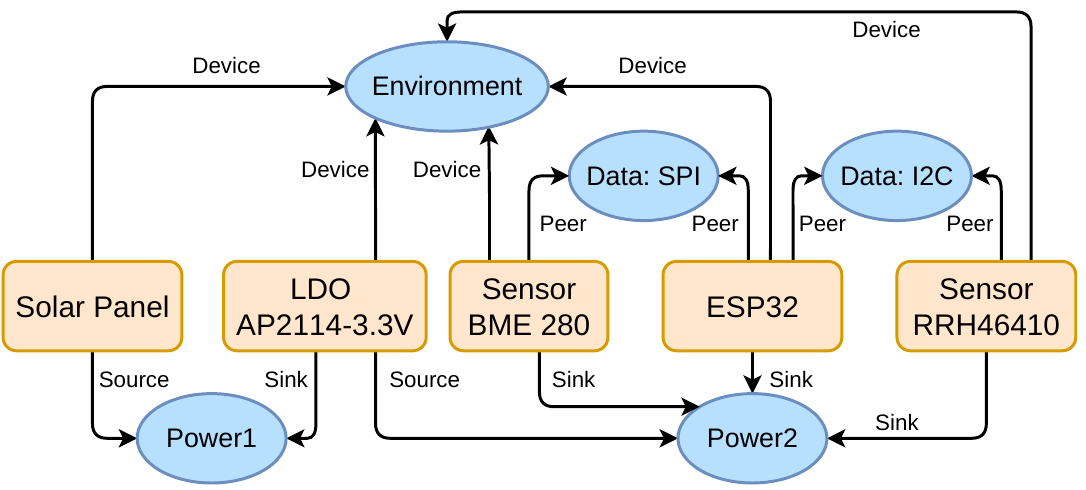}
    \caption{An example of the design graph with five devices in the system. Blue ellipses are netnodes and orange rectangles are device nodes. The design graph is generated from the high-level connectivity description: ``The AP2114-3.3 through VIN pin is powered by the solar panel through JST2.0. The ESP32 through 3V3 pin, the BME280 through its VDD pin, and RRH46410 through its VDD pin are powered by the AP2114-3.3 through VOUT pin. The BME280 transmits data to the ESP32 via SPI. The RRH46410 transmits data to the ESP32 via I2C.  All devices are physically in the same environment.''}
    \label{fig:DG-example}
\end{figure}

To enable scalable and connectivity-aware compatibility verification, we introduce the design graph (Component (b) in \autoref{fig:DG-pipeline}), called DG, to decompose system-level hardware compatibility analysis into connectivity-aware sub-tasks. By representing hardware connectivity as separable netnodes derived from the original connectivity, the design graph allows each compatibility verification task to consider only relevant component interactions, thereby reducing unnecessary context and mitigating LLM attention dilution. As the number of components and connections increases, the verification workload is distributed across modular and compact compatibility contexts with bounded information exchange, rather than requiring a single reasoning process over the entire hardware specification space, improving the scalability of LLM-guided reasoning. 
In other words, DEVICES scales by bounding the context required for each compatibility evaluation. Each verification task only considers the components and engineering properties associated with a specific netnode. As a result, its context size is primarily determined by the relevant component interactions rather than the entire hardware datasheet content. As hardware systems grow with more components and datasheets, DEVICES introduces additional modular verification tasks while maintaining compact reasoning contexts for each task, instead of continuously expanding a global context. In contrast, approaches that directly provide all datasheets to the LLM require the reasoning context to grow with the amount of available hardware information, increasing irrelevant information exposure and degrading reasoning performance as system complexity increases.

The DG contains three types of netnodes: power netnodes, data netnodes, and environment netnodes.

A \emph{power netnode} represents a power-supply domain shared by one or more devices. A device providing power to the domain is connected to the power netnode through an edge labeled \textit{source}, whereas a device consuming power from the domain is connected through an edge labeled \textit{sink}. This representation also distinguishes different power domains of the same device when they are supplied through different paths.

A \emph{data netnode} represents an inter-device communication channel or bus. Devices participating in the same communication channel are connected to a common data netnode through edges labeled \textit{peer}. For example, devices communicating over the same I$^2$C bus are connected to a common data netnode. The interface used by the devices, such as SPI, UART, or I$^2$C, is associated with the corresponding data netnode interaction and determines the applicable verification criteria.

An \emph{environment netnode} represents a shared physical operating environment. Devices that operate under the same environmental conditions are connected to the environment netnode through edges labeled \textit{device}. This representation allows environmental compatibility to be evaluated over the set of devices exposed to the same environment.

\autoref{fig:DG-example} illustrates an example DG generated from the high-level connectivity description specified in the caption. The resulting DG represents the power relationships through power netnodes, the communication relationships through data netnodes, and the shared operating conditions through an environment netnode. 

To construct the DG, the LLM receives the natural-language hardware component connectivity description and is instructed to instantiate the required device and netnodes and their corresponding edges. Rather than directly generating the complete graph, the LLM produces node and edge definitions in dedicated structured formats. A local deterministic script validates the generated structures and, when valid, constructs and caches the DG locally for further steps. This intermediate representation provides an explicit and inspectable representation of the system connectivity for subsequent property retrieval and compatibility evaluation.

\subsubsection{High-Level Component Connectivity Description}
The high-level connectivity description must provide sufficient information to instantiate the relevant netnodes and their relationships. For power netnodes, the description specifies the power pins and corresponding power domains of the devices, since a device may contain multiple power domains. For example, an EKG sensor may require separate digital-circuit and LED power domains, which may operate at different voltages and require different power sources. For data netnodes, the description specifies the communication interface used by each participating device, such as SPI, UART, or I$^2$C. For environment netnodes, the description identifies the devices that operate within the same physical environment.

We intend to minimize user effort in providing high-level connectivity descriptions. Nevertheless, these descriptions still require some hardware-specific information, such as power pins and power domains, which may require users to inspect relevant portions of datasheets, including pin functions and pin configurations. Although this effort is substantially lower than designing a detailed schematic, it remains manual. In ongoing work, we plan to develop an interactive LLM-assisted synthesis mechanism that helps users construct connectivity descriptions from the hardware knowledge graphs, further reducing the need for manual datasheet inspection and improving the overall degree of automation.

\subsection{Property Retrieval and Caching}

\label{subsec:property retrieval}

DEVICES retrieves the engineering properties required for compatibility evaluation from the hardware knowledge graph through a hierarchical, task-aware retrieval process (Component (c) in \autoref{fig:DG-pipeline}). For each device participating in a target netnode, the retrieval process is performed with the required property descriptions specified by each applicable verification criterion. The retrieved properties are then associated with the corresponding device and interactions in the design graph.

\subsubsection{Hierarchical and Task-Aware Retrieval}

Retrieving relevant engineering properties from the complete set of extracted properties of a hardware component remains challenging because a device may contain numerous properties distributed across different functional blocks and abstraction levels. Without hierarchical organization, the LLM must reason over all extracted properties to identify those relevant to the target compatibility analysis, resulting in unnecessary information processing, increased query overhead, and higher token costs. To address this challenge, the knowledge graph organizes properties hierarchically and enables a coarse-to-fine retrieval strategy that progressively narrows the search space. Specifically, the retrieval process first explores the block hierarchy of the target device and traverses block nodes to identify functional blocks that may contain properties required by the target verification task. Properties associated with the selected blocks are then retrieved together with properties from their ancestor nodes along the paths to the root, as higher-level nodes may contain general characteristics applicable to their descendant functional blocks.

The candidate properties obtained from the relevant hierarchy are subsequently provided to the LLM together with the required property descriptions for the target netnode. The LLM determines whether each candidate property is relevant to a required property and assigns the property to the corresponding requirement. This two-stage retrieval process first narrows the search space according to the semantic hierarchy and then performs property-level relevance matching. This substantially reduces the number of candidate properties, on average by 70.4\% compared with evaluating all properties in the device knowledge graph. When a cache hit occurs (see \autoref{subsubsec:retrieval caching}), the relevant properties are retrieved directly from the cache rather than selected through LLM-based retrieval reasoning. Such cases therefore incur zero token cost and involve zero newly considered properties. Accordingly, when computing the average reduction in considered properties, we exclude cache-hit cases.

LLMs may inadvertently alter retrieved engineering properties when reproducing them as outputs. Such modifications can introduce unintended changes to numerical values, conditions, or descriptions and compromise the fidelity of the verification process. To avoid this problem, DEVICES does not require the LLM to regenerate property contents. Instead, the LLM generates structured operations that reference properties through their unique property node IDs in the knowledge graph. A local deterministic script then executes these operations on the original properties stored locally without modifying their contents, avoiding the need for LLMs to regenerate the properties as outputs and reducing the risk of inadvertent content changes. Furthermore, the predefined operation structure enables automatic validation of the LLM outputs, allowing generation errors such as invalid references or malformed operations to be detected before subsequent processing. The selected properties are subsequently retrieved directly from the locally stored knowledge graph rather than being regenerated by the LLM.

\subsubsection{Model-Aware Property Retrieval}

The retrieval process is also model-aware because a single datasheet may contain specifications for multiple device variants, with each property associated either with the complete device family or with a specific model. Without preserving these associations, LLMs may incorrectly combine specifications from different variants and generate invalid compatibility constraints. Therefore, DEVICES filters properties according to their model applicability and only retrieves specifications associated with the target model or the entire device family. The retrieval process therefore excludes properties associated exclusively with non-target models and retains properties applicable to the target model or to the entire device series. For example, an AP2114 datasheet specifies electrical characteristics shared across the series as well as characteristics associated with individual voltage variants. Model-aware retrieval ensures that properties belonging to unrelated variants are not introduced into the compatibility evaluation context.

\subsubsection{Retrieval Caching}
\label{subsubsec:retrieval caching}

We use an automated caching mechanism to avoid repeated retrieval operations and reduce the cost of subsequent compatibility evaluations. Property retrieval can be token- and time-intensive because a device may contain hundreds of candidate properties that require semantic evaluation against the required properties. Although processing more candidates in a single query may reduce the number of LLM calls, it significantly increases the context size and may aggravate information omission. Therefore, DEVICES limits the number of candidate properties provided in each retrieval query to 20, maintaining a compact context while preserving retrieval accuracy.

The key observation enabling caching is that property retrieval is independent of the connectivity topology once the target verification domain is determined. Specifically, the relevant properties of a device depend on the device model and the verification domain, such as its communication interface, power role, or environmental requirement, rather than the identities of other connected devices. Therefore, after retrieving the relevant properties for a given device-domain context, DEVICES stores and reuses them whenever the same context appears in subsequent compatibility evaluations.

The cache is populated automatically from retrieval results generated by the pipeline rather than from manually specified or hardcoded device properties. When a matching cache entry is available, DEVICES bypasses the LLM retrieval process for that device-domain context, reducing its LLM token consumption to zero.

\subsection{Verification Script Generation and Constraint Evaluation}
\label{subsec: verification script generation and constraint evaluation}
For each netnode, DEVICES provides the LLM with the retrieved engineering properties of the participating devices and instructs it to generate a Python script that evaluates the compatibility constraints defined by the applicable verification criteria (Component (d) in \autoref{fig:DG-pipeline}). The LLM is not used to perform the final numerical computations or comparisons: these are not strengths of LLMs. Instead, all numerical operations, including unit conversion, range comparison, and mathematical computation, are executed deterministically by the generated Python scripts. The LLM is instructed to include explanatory comments documenting the implementation logic for each verification procedure.

\subsubsection{Task-Aware Context Construction}

Within each generated script prompt, the required properties are placed directly adjacent to the corresponding verification instructions for each participating device. Although the same property may appear under multiple instructions, this controlled redundancy improves the association between instructions and their required properties. This organization exploits the tendency of LLMs to use information more reliably when relevant instructions are located close together in the context. In our experiments, placing all properties in a single shared context without duplication results in more frequent property omission, whereas associating each instruction directly with its required properties improves instruction following and property use.

\subsubsection{Data and Environmental Compatibility}

Data netnode compatibility evaluation considers interface-mode compatibility, logic-level compatibility, and clock-frequency compatibility according to the applicable communication-interface verification criteria. Specifically, DEVICES evaluates whether participating devices support compatible communication modes, whether their logic voltage levels are mutually compatible, and whether their supported clock-frequency ranges permit a common operating frequency. Interface-specific electrical requirements are also considered when applicable; for example, I$^2$C compatibility additionally requires verification of the devices' sink-current capabilities.

For environment netnodes, compatibility is evaluated by determining whether the operating ranges of all participating devices have a non-empty intersection for all modeled environmental parameters, including temperature, humidity, and pressure. This makes it easy for the designer to determine whether the design is suitable for the intended operating environment. Data and environment netnodes are evaluated independently because their compatibility constraints do not propagate across different netnodes.

\subsubsection{Power Compatibility and Cascaded Power Evaluation}

Power compatibility considers input- and output-voltage compatibility, power capacity, and power-delivery relationships. Unlike data and environment compatibility, power constraints can propagate across transitively connected power netnodes. A device that consumes power from one power netnode may itself provide power to downstream devices through another power netnode, requiring the power demand of downstream devices to be propagated upstream through the power-delivery tree.

To provide a unified representation of power flow, we model a device as potentially acting as both a power sink and a power source, where either its input power or output power may be zero. In the current framework, only power regulators, such as linear regulators and motor drivers, are modeled as both sinks and sources. Other components, including sensors and MCUs, consume power from a dedicated source or regulator but do not supply power to downstream components.

The input power for device $D$ is modeled as follows:
\[
P_D = P_D^{\mathrm{self}} + \sum_{S \in \mathrm{sinks}} \frac{P_S}{\eta_{S-D}},
\]
where $P_D^{\mathrm{self}}$ is the device's own power consumption, $P_S$ is the power required by a downstream sink $S$, and $\eta_{S-D}$ is the power-transfer efficiency associated with the corresponding power-conversion path, i.e., from device S to device D. This model accounts for both direct and transitive power demands and considers power conversion losses.

In cascaded power delivery networks, power demand accumulates along the power path and must be propagated across connected power netnodes. However, DEVICES constructs contexts independently for individual netnodes to maintain compact and task-specific LLM inputs. Therefore, accumulated power demand from downstream netnodes must be explicitly conveyed to upstream evaluations.  

Rather than merging all devices and verification instructions from multiple cascaded power netnodes into a single LLM context, which would cause the context size to increase with the depth and breadth of the power tree, we propagate computed power demand between adjacent power netnodes. Specifically, the LLM is instructed to compute the aggregated downstream power demand for each power netnode and store the result in a dedicated output variable. A local deterministic script evaluates the generated program starting from leaf power netnodes, extracts the computed power variable, and provides it as an input property to the parent-node evaluation. This process recursively propagates accumulated power demand from leaf nodes toward the root while preserving modular LLM contexts for individual power netnodes.

For each power netnode, compatibility evaluation verifies that the maximum of peak power demands of its downstream sinks does not exceed the maximum output power capability of the supplying device. When a sink supports multiple operating modes, its maximum demand is taken over all supported modes. The resulting constraint is
\[
\sum_{S \in \mathrm{sinks}}
\frac{\max_{m \in \mathrm{modes}}\left\{P_{m,S}^{\mathrm{peak}}\right\}}
{\eta_S}
\leq P_D^{\mathrm{worst}},
\]
where $P_{m,S}^{\mathrm{peak}}$ is the peak power consumption of sink $S$ in operating mode $m$, $\eta_S$ is the corresponding power-transfer efficiency, and $P_D^{\mathrm{worst}}$ is the worst-case guaranteed output power capability of device $D$.

This provides a conservative estimate, as the computed power consumption may be higher than the actual power consumption of some systems.

\section{Evaluation}

\subsection{Evaluation Setup}

We evaluate DEVICES on seven embedded-system designs and 34 hardware datasheets (1210 pages in total). Multiple device variants described within the same datasheet are evaluated independently to assess model-aware knowledge extraction and retrieval, but are counted as a single datasheet. The seven designs cover three representative IoT application domains: environmental sensing, health monitoring, and location tracking.

The evaluation designs contain commonly used power sources, power-management components, and communication interfaces as well as commonly encountered environmental conditions. For communication compatibility, we evaluate embedded system serial interfaces including UART, SPI, and I$^2$C. For environmental compatibility, we evaluate temperature, humidity, and pressure. 

Several power modules, including solar panels, batteries, power adapters, and LDOs are evaluated. As explained in \autoref{sec:introduction}, switching regulators, such as buck converters, are not currently supported because their efficiency is often provided as curves in figures, while figure understanding is not yet supported by DEVICES. These categories cover common power, communication, and environmental requirements encountered in the evaluated embedded-system designs. 

We evaluate DEVICES from three perspectives: \emph{compatibility verification accuracy}, \emph{context size}, and \emph{property faithfulness}.

\textbf{Compatibility verification accuracy} measures the proportion of correctly evaluated compatibility constraints among all evaluated constraints. A constraint is considered correctly evaluated only if the generated evaluation uses the correct engineering properties, follows the required evaluation procedure, and reaches the correct compatibility conclusion. An error in any of these components is counted as an incorrect evaluation. For example, even if two devices are correctly concluded to have compatible logic-high voltage levels, using an incorrect voltage specification to reach this conclusion is considered an incorrect evaluation. This is a very strict evaluation metric that captures the end-to-end correctness of compatibility verification rather than only the correctness of the final compatibility decision.

\textbf{Context size} measures the number of input tokens provided to the LLM for generating each compatibility evaluation script. It characterizes the amount of information exposed to the model during evaluation and serves as an indicator of long-context overhead and the potential for information dilution. Furthermore, the growth of context size with increasing system complexity reflects the scalability of the evaluated approaches. Specifically, as the number of components and interactions increases, we compare how the required context size grows for DEVICES and the two baselines. For the same compatibility evaluation task, a smaller context indicates that the model processes less irrelevant or redundant information, enabling a more focused and controlled evaluation context. Note that the reported context token counts of DEVICES represent the summation of the input tokens across all netnodes in the DG of each design during the final Verification Script Generation step introduced in \autoref{subsec: verification script generation and constraint evaluation}. Therefore, the context size used for generating the verification script of each individual netnode is smaller than the aggregated values reported here.

\textbf{Property faithfulness} measures whether the engineering properties used by the generated evaluation scripts are faithful to the properties explicitly provided in the input context. A property is considered faithful when the script uses the provided property without modifying, inventing, or introducing unsupported information. This metric evaluates whether the generated evaluation remains grounded in the retrieved knowledge rather than relying on information implicitly regenerated or hallucinated by the LLM.

\subsection{Ground Truth}

We establish the ground truth through manual examination of the corresponding hardware datasheets and high-level connectivity descriptions. For each design, we identify the engineering properties required by the applicable compatibility constraints, formulate the corresponding verification procedures, and manually determine the expected compatibility conclusions. The ground truth is independently evaluated to ensure that the evaluated properties, verification procedures, and conclusions accurately reflect the specifications provided by the component manufacturers.

\subsection{Baseline}

We compare DEVICES against two one-shot prompting baselines. The first baseline represents the common workflow in which users directly provide documents and ad hoc queries to an LLM without manually constructing detailed verification criteria. For each design, all relevant hardware datasheets are provided to the LLM together with the same high-level connectivity description used by DEVICES. The LLM is then instructed to directly evaluate hardware compatibility based on the provided information. This baseline reflects a practical usage scenario of current LLM-assisted document analysis, where users rely on ad hoc prompts rather than domain-specific verification procedures.

The second baseline is designed as an ablation study to evaluate the individual impacts of two of our key contributions: verification criteria and task-aware context refinement. This baseline uses the same one-shot document prompting setting as the first baseline but is also provided with the verification criteria used in DEVICES, including the verification procedures, required engineering properties, and compatibility rules. However, unlike DEVICES, it does not perform structured property retrieval or compact task-specific context construction. Comparing this baseline with DEVICES reveals the impact of context refinement and selective property exposure beyond the benefits provided by explicit verification instructions.

We intentionally do not provide the first baseline with manually constructed verification criteria because such criteria require substantial hardware expertise and prompt engineering effort, which represents a core contribution of DEVICES. Including these specifications would result in a baseline that is not representative of typical LLM use.

\subsection{Results}
\subsubsection{Compatibility verification accuracy}
\label{subsubsec:compatibility checking accuracy}
\begin{figure}[!t]
    \centering
    \includegraphics[width=\linewidth]{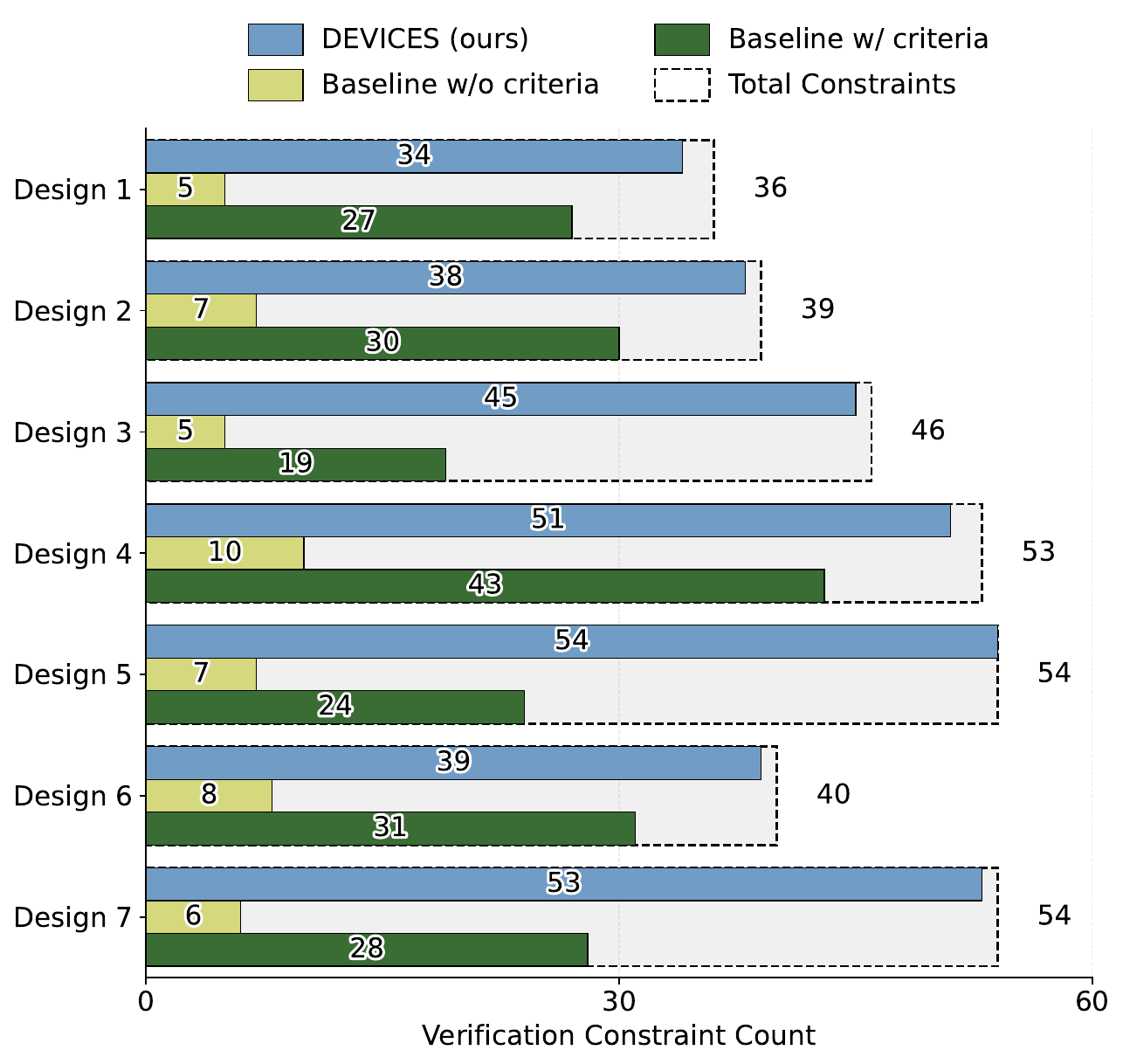}
    \caption{Compatibility verification accuracy of DEVICES and two baselines with and without verification criteria. Values inside the bars indicate the number of correctly evaluated constraints. Constraint evaluation is considered correct only if it uses the correct engineering properties, follows the required evaluation procedure, and reaches the correct compatibility conclusion; any violation is counted as failure. DEVICES achieves 97.5\% accuracy, compared with 14.9\% for the baseline without verification criteria and 62.7\% for the baseline with verification criteria. These results demonstrate the importance of criteria and task-aware context refinement for early-stage, datasheet-aware compatibility verification.
    }
    \label{fig:accuracy}
\end{figure}

\autoref{fig:accuracy} presents the compatibility verification accuracy of DEVICES and two baselines across seven designs. Each design contains at least five distinct hardware components. DEVICES achieves an overall accuracy of 97.5\%, while the one-shot document prompting baseline without verification criteria achieves only 14.9\% accuracy. When enhanced with the verification criteria, the one-shot baseline improves to 62.7\% accuracy.

For the baseline without verification criteria, we observe that the generated evaluation scripts frequently omit required verification criteria and engineering properties. Moreover, the evaluated criterion are inconsistent across runs, where the same compatibility requirement may be verified in one execution but ignored in another. These results indicate that directly relying on the evaluated LLM to infer complete compatibility requirements from lengthy hardware documentation is insufficient for reliable compatibility verification. Explicitly defining the required verification criteria is therefore necessary to guide LLMs toward consistent verification behavior.

After incorporating the verification criteria, the one-shot baseline achieves a substantial accuracy improvement, demonstrating that explicit verification instructions are essential for reliable LLM-based compatibility evaluation. We attribute this improvement to two factors. First, the criteria provide unambiguous evaluation procedures that reduce the ambiguity of compatibility verification. Second, the explicitly specified required properties provide semantic anchors that help the LLM identify relevant information from the documents and reduce property omission. The generated scripts further confirm that the LLM follows the provided verification procedures more consistently.

However, despite the improvement, the criteria-enhanced one-shot baseline still makes substantial errors. Through manual inspection of the generated scripts, we found that the LLM occasionally combines specifications from incompatible operating conditions or device variants, i.e., using the sink current specification of five-volt-tolerant Fast-Mode Plus I$^2$C pins while using logic-level voltage specifications of \qty{3.3}{V} pins, resulting in inconsistent compatibility verification. We attribute these errors to the long and sparse document contexts provided in the one-shot setting, where related but condition-dependent specifications are distributed across datasheets. This observation demonstrates that explicit verification criteria alone are insufficient; task-aware context construction and selective property exposure are also necessary for reliable compatibility verification.

\subsubsection{Context size}

\begin{figure}[!t]
    \centering
    \includegraphics[width=\linewidth]{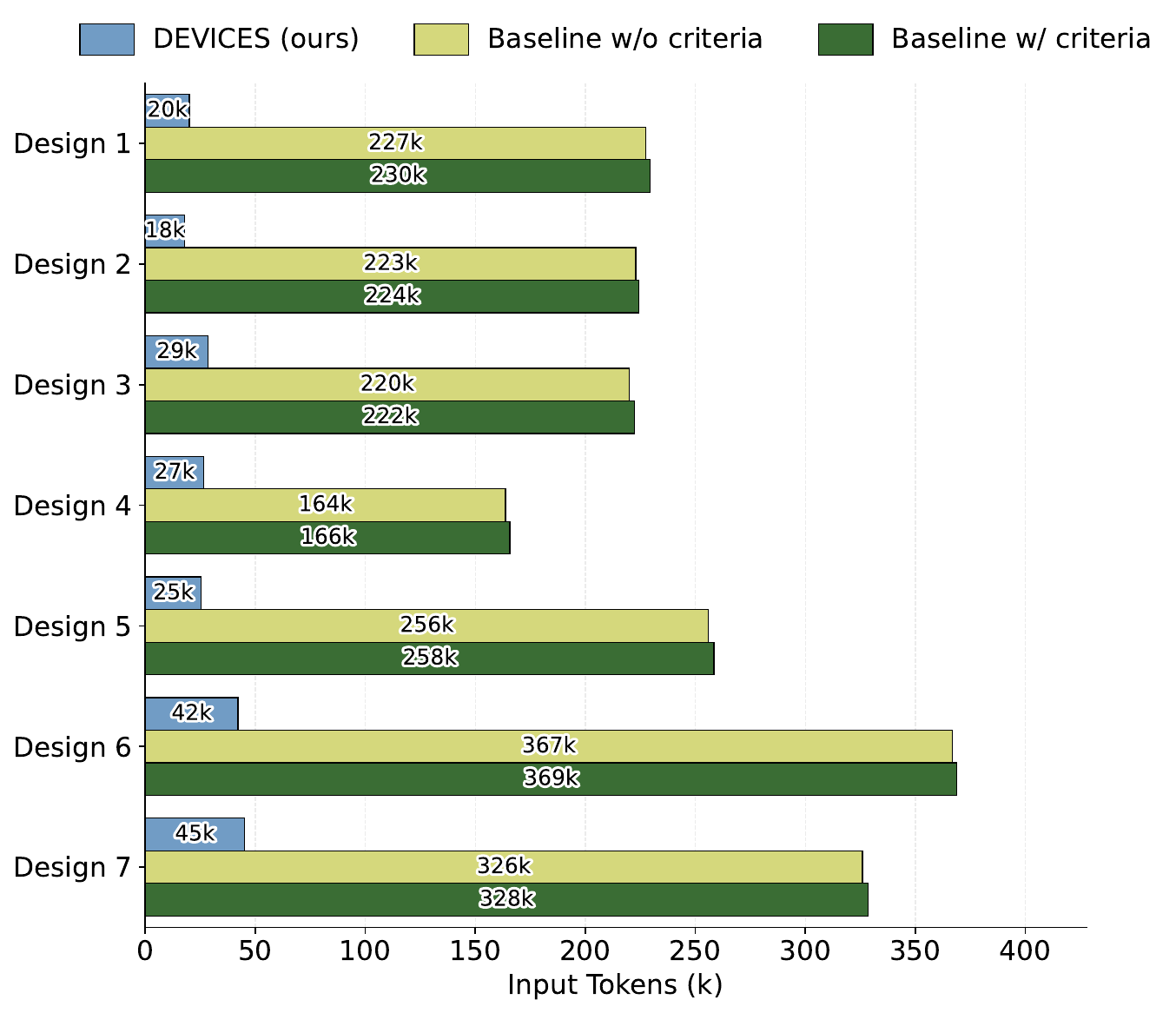}
    \caption{Input tokens of DEVICES and the two baselines with and without verification criteria. Input-token count measures the context size provided to the LLM. We use context size as an indicator of the amount of information the LLM must process, motivated by our evaluation results and prior studies showing that increasing context size degrades LLM performance, particularly when relevant information is surrounded by irrelevant content. For DEVICES, the reported value is the sum of input tokens across all netnode queries of each design; therefore, the context size of each individual query is substantially smaller than the reported total. The maximum context size is approximately 12.6k input tokens and the average context size per query is approximately 4.7k input tokens. The results also demonstrate the scalability of our modular context construction: as the design grows in complexity, DEVICES processes each netnode using a task-specific context rather than repeatedly inputting the complete datasheets to the LLM. Consequently, its per-query context remains substantially smaller than that of both one-shot baselines, allowing the LLM to focus on the relevant properties and verification instructions.}
    \label{fig:tokens}
\end{figure}

The input context sizes, measured by the number of input tokens, are shown in \autoref{fig:tokens}. Although context size provides an indirect measure of the information exposed to the LLM, our ablation studies in \autoref{subsubsec:compatibility checking accuracy} demonstrate its impact on compatibility verification accuracy. As shown in \autoref{fig:tokens}, DEVICES substantially reduces the input context size compared with the one-shot approaches. Importantly, the reported input token count for DEVICES is the aggregate across all netnodes within each design; thus, the context required for an individual verification query is substantially smaller than the reported aggregate and is also smaller than that of the one-shot approaches. By restricting each verification query to task-relevant component interactions and engineering properties, DEVICES reduces the amount of irrelevant information exposed to the LLM, providing a more focused reasoning context. More importantly, the results demonstrate the scalability advantage of the proposed decomposition: the context size of each individual query is primarily determined by the involved connectivity relationship and the associated properties, rather than by the total amount of hardware specification information in the design. Consequently, increasing the number of components and datasheets primarily introduces additional modular verification queries instead of proportionally expanding the context of each query.

\subsubsection{Property faithfulness}

During the manual evaluation of compatibility verification accuracy, we determine whether the variables used in the generated scripts are consistent with the corresponding datasheets and whether they are directly derived from the properties provided in the input contexts. A verification constraint is considered incorrect if it uses an incorrect property, including a property with an incorrect value or target, an incorrect operating condition, an incorrect device variant, or information derived from external knowledge that does not appear in either the datasheets or the provided contexts. Under this definition, we find that DEVICES consistently use only properties available in the provided contexts and do not introduce external properties. The two baselines likewise adhere to this property-provenance requirement. Therefore, the observed accuracy differences primarily reflect differences in property selection, condition consistency, and evaluation procedures rather than the use of external information.

\section{Discussion}
\label{sec: discussion}

The continuous evolution of LLMs raises the question of whether improvements in model capability, particularly larger context windows and stronger reasoning ability, may eventually eliminate the need for context refinement and task decomposition. We argue that such improvements will not diminish the value of structured LLM-guided hardware verification frameworks, such as DEVICES.

First, improved LLM capabilities are expected to enable the analysis of increasingly complex systems rather than simply solving existing problems with larger contexts. As embedded systems evolve from individual boards toward highly integrated cyber-physical systems with increasing numbers of components, interfaces, and constraints, the amount of engineering information involved in compatibility analysis will continue to grow. Therefore, the challenge shifts from whether an LLM can process a fixed amount of information to how complex engineering problems can be decomposed into manageable reasoning tasks. By leveraging design graphs and domain-oriented verification criteria, DEVICES decomposes system-level compatibility evaluation into localized verification tasks and constructs compact contexts containing only the properties required for each analysis. This allows the verification workload to scale with relevant component interactions rather than requiring a single reasoning process over the entire hardware specification space.

Second, context construction remains valuable even if future LLMs can process substantially larger contexts. The purpose of context construction is not solely to overcome limited context windows, but also to provide task-specific evidence that improves transparency, reliability, and interpretability. By explicitly selecting compatibility-relevant properties and excluding unrelated information, DEVICES reduces unnecessary information processing, enables traceable verification decisions, and ensures that compatibility evaluation follows well-defined engineering criteria.

Consequently, future improvements in LLM capabilities can further enhance each localized reasoning task within DEVICES rather than replace the need for such a structured verification architecture. More capable LLMs can improve individual components of DEVICES, including hardware knowledge extraction, relevant property identification, and verification procedure generation, while the overall architecture continues to provide the necessary organization for scalable and reliable evaluation. DEVICES is not merely a workaround for current LLM limitations, but an extensible architecture that can leverage future LLM advances to enable increasingly scalable and reliable hardware compatibility evaluation.

\section{Conclusion}
This paper described DEVICES, an LLM-driven framework for early-stage, pre-schematic hardware compatibility verification based on datasheets and high-level component connectivity descriptions. DEVICES combines compact, task-oriented context construction and verification criteria to guide LLMs toward relevant engineering information and consistent, complete compatibility evaluation. By decomposing system-level verification into smaller tasks and preserving intermediate results, DEVICES improves transparency and interpretability while enabling the verification workload to scale with relevant component interactions rather than the total hardware specification content. Across seven embedded system designs, DEVICES achieved 97.5\% compatibility-checking accuracy and substantially reduced the input context size by 8.6$\times$ compared with one-shot full-PDF prompting. These results demonstrate the feasibility of using LLMs for automated, specification-based compatibility verification before detailed schematic design and simulation, providing an effective, scalable early-stage screening for embedded-system hardware integration. Additionally, they demonstrate the need for, and substantial benefits of, modular task decomposition, formalized verification criteria, and task-aware compact context construction.

\newpage

\bibliographystyle{IEEEtran}
\bibliography{reference}
\end{document}